# Stylistic Variation in Social Media Part-of-Speech Tagging

**Murali Raghu Babu Balusu** and **Taha Merghani** and **Jacob Eisenstein**
School of Interactive Computing
Georgia Institute of Technology
Atlanta, GA, USA
{b.murali, tmerghani3, jacobe}@gatech.edu

## Abstract

Social media features substantial stylistic variation, raising new challenges for syntactic analysis of online writing. However, this variation is often aligned with author attributes such as age, gender, and geography, as well as more readily-available social network metadata. In this paper, we report new evidence on the link between language and social networks in the task of part-of-speech tagging. We find that tagger error rates are correlated with network structure, with high accuracy in some parts of the network, and lower accuracy elsewhere. As a result, tagger accuracy depends on training from a balanced sample of the network, rather than training on texts from a narrow subcommunity. We also describe our attempts to add robustness to stylistic variation, by building a mixture-of-experts model in which each expert is associated with a region of the social network. While prior work found that similar approaches yield performance improvements in sentiment analysis and entity linking, we were unable to obtain performance improvements in part-of-speech tagging, despite strong evidence for the link between part-of-speech error rates and social network structure.

## 1 Introduction

Social media feature greater diversity than the formal genres that constitute classic datasets such as the Penn Treebank (Marcus et al., 1993) and the Brown Corpus (Francis and Kucera, 1982): there are more authors, more *kinds* of authors, more varied communicative settings, fewer rules, and more stylistic variation (Baldwin et al., 2013; Eisenstein, 2013). Previous work has demonstrated precipitous declines in the performance of state-of-the-art systems for core tasks such as part-of-speech tagging (Gimpel et al., 2011) and named-entity recognition (Ritter et al., 2010) when these systems are applied to social media text, and stylistic diversity seems the likely culprit. However, we still lack quantitative evidence of the role played by language variation in the performance of NLP systems in social media, and existing solutions to this problem are piecemeal at best. In this paper, we attempt to address both issues: we quantify the impact of one form of sociolinguistic variation on part-of-speech tagging accuracy, and we design a model that attempts to adapt to this variation.

Our contribution focuses on the impact of language variation that is aligned with one or more *social networks* among authors on the microblogging platform Twitter. We choose Twitter because language styles in this platform are particularly diverse (Eisenstein et al., 2010), and because moderately large labeled datasets are available (Gimpel et al., 2011; Owoputi et al., 2013). We choose social networks for several reasons. First, they can readily be obtained from both metadata and behavioral traces on multiple social media platforms (Huberman et al., 2008). Second, social networks are strongly correlated with "demographic" author-level variables such as age (Rosenthal and McKeown, 2011), gender (Eckert and McConnell-Ginet, 2003), race (Green, 2002), and geography (Trudgill, 1974), thanks to the phenomenon of *homophily*, also known as *assortative mixing* (McPherson et al., 2001; Al Zamal et al., 2012). These demographic variables are in turn closely linked to language variation in American English (Wolfram and Schilling-Estes, 2005), and have been shown to improve some document classification tasks (Hovy, 2015). Third, there is growing evidence of the strong relationship between social network structures and language variation, even beyond the extent to which the social network acts as a proxy for demographic attributes (Milroy, 1991; Dodsworth, 2017).

To measure the impact of socially-linked language variation, we focus on part-of-speech tagging, a fundamental task for syntactic analysis. First, we measure the extent to which tagger performance is correlated with network structure, finding that tagger performance on friends is significantly more correlated than would be expected by chance. We then design alternative training and test splits that are aligned with network structure, and find that test set performance decreases in this scenario, which corresponds to domain adaptation across social network communities. This speaks to the importance of covering all relevant social network communities in training data.

We then consider how to address the problem of language variation, by building social awareness into a recurrent neural tagging model. Our modeling approach is inspired by Yang and Eisenstein (2017), who train a mixture-of-experts for sentiment analysis, where the expert weights are computed from social network node embeddings. But while prior work demonstrated improvements in sentiment analysis and information extraction (Yang et al., 2016), this approach does not yield any gains on part-of-speech tagging. We conclude the paper by briefly considering possible reasons for this discrepancy, and propose approaches for future work in social adaptation of syntactic analysis.[1]

## 2 Data

We use the corrected[2] OCT27 dataset from Gimpel et al. (2011) and Owoputi et al. (2013) as our training set, which contains part-of-speech annotations for 1,827 tweets sampled from Oct 27-28, 2010. We use the train and dev splits of OCT27 as our training dataset and the test split of OCT27 dataset as our validation dataset. The DAILY547 dataset from Owoputi et al. (2013) which has 547 tweets is used for evaluation. Table 2 specifies the number of tweets and tokens in each dataset. The tagset for this dataset is explained in Owoputi et al. (2013); it differs significantly from the Penn Treebank and Universal Dependencies tagsets.

In September 2017, we extracted author IDs for each of the tweets and constructed three author social networks based on the follow, mention, and retweet relations between the authors in the dataset, which we refer to as *follow*, *mention* and *retweet* networks in Table 2. Specifically, we use the Twitter API to crawl the friends of the OCT27 and DAILY547 users (individuals that they follow) and the most recent 3,200 tweets in their timelines. The mention and retweet links are then extracted from the tweet text and metadata. Table 2 specifies the total number of authors (whose tweets exist in our dataset) in each network, the total number of nodes and the total number of relations among these nodes. We treat all social networks as undirected graphs, where two users are socially connected if there exists at least one social relation between them. Several authors of the tweets can no longer be queried from Twitter, possibly because their accounts have been deleted. They are not included in the network, but their tweets are still used for training and evaluation.

---

[1] Code for rerunning the experiments is available here: https://github.com/bmurali1994/socialnets_postagging

[2] Owoputi et al. corrected inconsistencies in the ground labeling of *that/this* in 100 (about 0.4%) total labels.

| Dataset | #Msg. | #Tok. |
|---|---|---|
| OCT27 | 1,827 | 26,594 |
| DAILY547 | 547 | 7,707 |

Table 1: Annotated datasets: number of messages and tokens

| Network | #Authors | #Nodes | #Edges |
|---|---|---|---|
| Follow | 1,280 | 905,751 | 1,239,358 |
| Mention | 1,217 | 384,190 | 623,754 |
| Retweet | 1,154 | 182,390 | 314,381 |

Table 2: Statistics for each social network

## 3 Linguistic Homophily

The hypothesis of *linguistic homophily* is that socially connected individuals tend to use language similarly, as compared to randomly selected pairs of individuals who are not socially connected (Yang and Eisenstein, 2017). We now describe two pilot studies that test this hypothesis.

### 3.1 Assortativity

We test whether errors in POS tagging are *assortative* on the social networks defined in the previous section: that is, if two individuals $(i, j)$ are connected in the network, then a model's error on the tweets of author $i$ suggests that the errors on the tweets of author $j$ are more likely. To measure assortativity, we compute the average difference in the tagger's per-token accuracy on tweets for au-

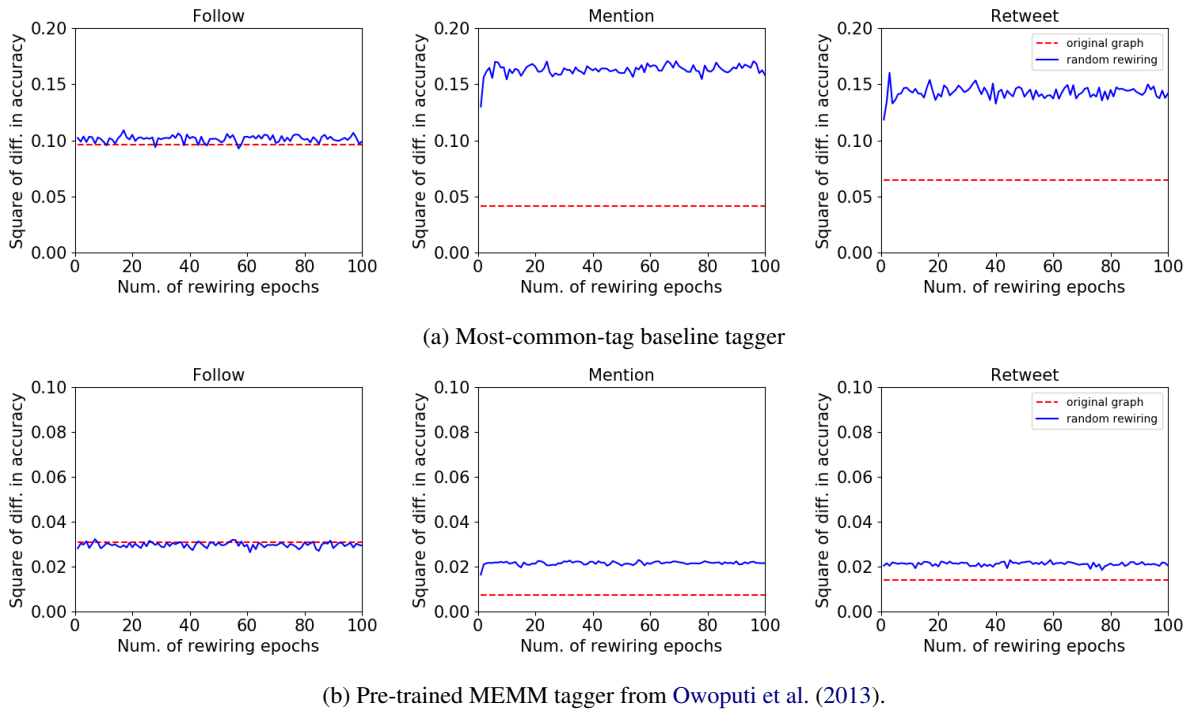

(a) Most-common-tag baseline tagger

(b) Pre-trained MEMM tagger from Owoputi et al. (2013).

Figure 1: Average of the squared difference in tagging accuracy on observed (red) and randomized networks (blue).

thors $i$ and $j$, averaged over all connected pairs in the network. This measures whether classification errors are related on the network structure.

We compare the observed assortativity against the assortativity in a network that has been randomly rewired. Each rewiring epoch involves a number of random rewiring operations equal to the total number of edges in the network. The edges are randomly selected, so a given edge may not be rewired in each epoch; furthermore, the degree of each node is preserved throughout. If the squared difference in accuracy is lower for the observed networks than for their rewired counterparts, this would indicate that tagger accuracy is correlated with network structure. Figure 2 explains the metric and rewiring briefly through an example.

We compute the assortativity for three taggers:

- We first use a naïve tagger, which predicts the most common tag seen during training if the word exists in the vocabulary, and otherwise predicts the the most common tag for an unseen word. Preprocessing of each tweet involves lowercasing, normalizing all @-mentions to $\langle @MENTION \rangle$, and normalizing URLs and email addresses to a common token (e.g. $http://bit.ly/dP8rR8 \Rightarrow \langle URL \rangle$).

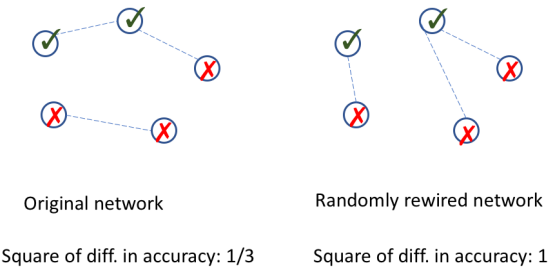

Figure 2: Toy example: differences in tagging accuracy on original and randomly-rewired network.

- We train a lexical, feature-rich CRF model. Lexical features in the CRF model include the word, previous two words, next two words, prefixes and suffixes of the previous two, current and next two words, and flags for special characters like hyphen, at-mention, hashtag, hyphen and digits in the current word.

- Finally, we repeat these experiments with the pretrained maximum entropy Markov model (MEMM) tagger from (Owoputi et al., 2013), trained on OCT27 tweets.

Figure 1 shows the results for the naïve tagger and the MEMM tagger; the results were similar for the CRF were similar. Tagger accuracy is well correlated with network structure in the mention and retweet graphs, consistent with the hypothesis of linguistic homophily. These findings support prior work suggesting that "behavioral" social networks such as mentions and retweets are more meaningful than "articulated" networks like the follower graph (Huberman et al., 2008; Puniyani et al., 2010).

## 3.2 Clustering

Next, we examine whether linguistic homophily can lead to mismatches between the test and training data. We embed each author's social network position into a vector representation of dimension $D_v$, using the LINE method for social network node embedding (Tang et al., 2015). These embeddings are obtained solely from the social network, and not from the text.

We obtain $D_v = 50$-dimensional node embeddings, and apply k-means clustering (Hartigan and Wong, 1979) to obtain two sets of authors (train and test). By design, the training and test sets will be in different regions of the network, so training and test authors will be unlikely to be socially connected. We then train the lexical CRF tagger on the training set, and apply it to the test set. The same setup is then applied to a randomly-selected training/test split, in which the social network structure is ignored. This comparison is illustrated in Figure 3. We repeat this experiment for 10 times for all three social networks: follow, mention and retweet.

The theory of linguistic homophily implies that the test set performance should be *worse* in the case that the test set and training sets are drawn from different parts of the network, since the linguistic style in the training set will not match the test data. In contrast, when the training and test sets are drawn in a manner that is agnostic to network structure, the training and test sets are expected to be more linguistically similar, and therefore, test set performance should be better. As shown in Table 3, the results support the theory: predictive accuracy is higher when the test and training sets are not drawn from different parts of the network.

## 4 Adapting to socially-linked variation

In this section, we describe a neural network method that leverages social network informa-tion to improve part-of-speech tagging. We employ the *Social Attention* neural network architecture, where the system prediction is the weighted combination of the outputs of several basis models (Yang and Eisenstein, 2017). We encourage each basis model to focus on a local region of the social network, so that classification on socially connected individuals employs similar model combinations. This allows sharing of strength for some similar properties between these network components.

| Network | Network clusters | Random |
|---|---|---|
| Follow | 82.01% | 83.83% |
| Mention | 81.40% | 83.07% |
| Retweet | 81.01% | 83.52% |

Table 3: Comparison of tagger accuracy using network-based and random training/test splits

In this architecture, each prediction is the weighted combination of the outputs of several basis models. Given a set of labeled instances $\{x_i, y_i\}$ and authors $\{a_i\}$, the goal of personalized probabilistic classification is to estimate a conditional label distribution $p(y \mid x, a)$. We condition on the author $a$ by modeling the conditional label distribution as a mixture over the posterior distributions of $K$ basis taggers,

$$p(y \mid x, a) = \sum_{k=1}^{K} \pi_{a,k} \times p_k(y \mid x) \quad (1)$$

The basis taggers $p_k(y \mid x)$ can be arbitrary conditional distributions. We use a hierarchical recurrent neural network model, in addition to a tag dictionary and Brown cluster surface features (Brown et al., 1992), which we describe in more detail in § 4.2. The component weighting distribution $\pi_{a,k}$ is conditioned on the social network $G$, and functions as an attentional mechanism, described in § 4.1. The main idea is that for a pair of authors $a_i$ and $a_j$ who are nearby in the social network $G$, the prediction rules should behave similarly if the attentional distributions are similar, i.e., $\pi_{a_i,k} \approx \pi_{a_j,k}$. If we have labeled training data for $a_i$ and wish to make predictions on author $a_j$, some of the personalization from $a_i$ will be shared by $a_j$. The overall classification approach can be viewed as a mixture of experts (Jacobs et al., 1991), leveraging the social network

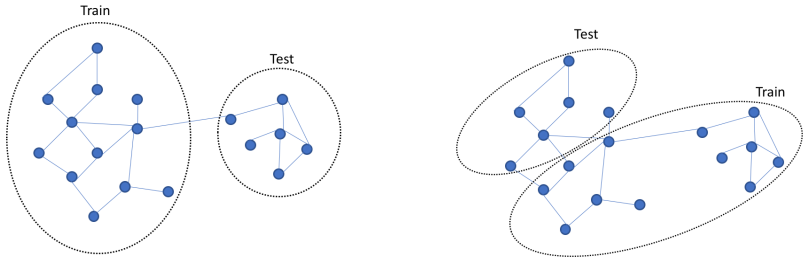

Figure 3: (left) Network-aligned train/test split and (right) random train/test split

as side information to choose the distribution over experts for each author.

### 4.1 Social Attention Model

The goal of the social attention model is to assign similar basis weights to authors who are nearby in the social network $G$. We operationalize social proximity by embedding each author's social network position into a vector representation, again using the LINE method for node embedding (Tang et al., 2015). The resulting embeddings $v_a$ are treated as fixed parameters in a probabilistic model over edges in the social network. These embeddings are learned solely from the social network $G$, without leveraging any textual information. The attentional weights are then computed from the embeddings using a softmax layer,

$$\pi_{a,k} = \frac{\exp(\boldsymbol{\phi}_k \cdot \boldsymbol{v}_a + b_k)}{\sum_{k'}^{K} \exp(\boldsymbol{\phi}_{k'} \cdot \boldsymbol{v}_a + b_{k'})}. \quad (2)$$

The parameters $\boldsymbol{\phi}_k$ and $b_k$ are learned in the model. We observed that almost 50% of the authors in our dataset do not appear in any social network. For all these authors, we use the same embedding $\boldsymbol{v}'$ to let the model learn the proportion weight of the individual basis models in the ensemble. This embedding $\boldsymbol{v}'$ is also learned as a parameter in the model. We have also tried computing the attentional weights using a sigmoid function,

$$\pi_{a,k} = \sigma(\boldsymbol{\phi}_k \cdot \boldsymbol{v}_a + b_k), \quad (3)$$

so that $\boldsymbol{\pi}_a$ is not normalized, but the results were quite similar.

### 4.2 Modeling Surface Features

We use surface-level features in addition to the basis models to improve the performance of our model closer to the state-of-the-art results. Specifically, we use the tag dictionary features and the Brown cluster features as described by Gimpel et al. (2011).

Since Brown clusters are hierarchical in a binary tree, each word is associated with a tree path represented as a bitstring with length $\leq 16$; we use prefixes of the bitstring as features (for all prefix lengths $\in \{2, 4, 6, ..., 16\}$). Concatenating the Brown cluster features of the previous and next token along with the current token helped improve the performance of the baseline model.

We also used the tag dictionary features from Gimpel et al. (2011), by adding features for a word's most frequent part-of-speech tags from Penn Treebank and Universal Dependencies. This also helped improve the performance of the baseline model. We found these surface features to be vital. Nonetheless, we were not able to match the performance of the state-of-the-art systems.

### 4.3 POS tagging with Hierarchical LSTMs

We next describe the baseline model: $p_k(\boldsymbol{y} \mid \boldsymbol{x})$. The baseline model is a word-level bi-LSTM, with a character-level bi-LSTM to compute the embeddings of the words (Ling et al., 2015). In addition to the embeddings from the character level bi-LSTM, we also learn the word embeddings which are initialized randomly and also use fixed pretrained GloVe Twitter (Pennington et al., 2014) embeddings for the word-level bi-LSTM. The final input to the word-level LSTM is the concatenation of the embedding from the character level, learned word embedding and the fixed pretrained word embedding. The final hidden state for each word $\boldsymbol{h}_i$ is obtained and concatenated with the surface features for each word $\boldsymbol{s}_i^k$, and the result is passed through a fully connected neural network, giving a latent representation $\boldsymbol{r}_i^k$. The conditional probability is then computed as,

$$p_k(y_i = t \mid x_i) = \frac{\exp(\boldsymbol{\beta}_t \cdot \boldsymbol{r}_i^k + c_t)}{\sum_{t'} \exp(\boldsymbol{\beta}_{t'} \cdot \boldsymbol{r}_i^k + c_{t'})}. \quad (4)$$

## 4.4 Loss Function and Training

We train the ensemble model by minimizing the negative log likelihood of the tags for all the tokens in all the tweets in the training dataset.

**Alternative objectives** We have also tried training the model using a hinge loss, but the results were similar and hence excluded in the paper. We also explored a variational autoencoder (VAE) framework (Kingma and Welling, 2014), in which the node embeddings were modeled with a latent vector $z$, which was used both to control the mixture weights $\pi_k$, and to reconstruct the node embeddings. Again, results were similar to those obtained with the simpler negative log-likelihood objective.

**Training problems** One potential problem with this framework is that after initialization, a small number of basis models may claim most of the mixture weights for all the users, while other basis models are inactive. This can occur because some basis models may be initialized with parameters that are globally superior. As a result, the "dead" basis models will receive near-zero gradient updates, and therefore can never improve. Careful initialization of the parameters $\phi_k$ and $b_k$ and using L2-regularization parameters of the model helped mitigate the issue to some extent. Using the attentional weights computed using the sigmoid function as described in Equation 3 does not have this problem, but the final evaluation results were quite similar to the model with attentional weights computed using softmax as mentioned in Equation 2.

## 5 Experiments

Our evaluation focuses on the DAILY547 dataset (Owoputi et al., 2013). We train our system on the train and dev splits of the OCT27 dataset (Gimpel et al., 2011) and use the test split of OCT27 as our validation dataset and evaluate on the DAILY547 dataset. Accuracy of the tokens is our evaluation metric for the model. We compare our results to our baseline model and the state of the art results on the Twitter OCT27+Daily547 dataset.

### 5.1 Experimental Settings

We use 100-dimensional pretrained Twitter GloVe embeddings (Pennington et al., 2014) which are trained on about two billion tweets. We use one-layer for both the character-level and the word-level bi-LSTM model with hidden state sizes of 50 and 150 dimensions respectively. The dimensions of character embeddings is set to be 30 and the learned word embeddings is 50. We use tanh activation functions all throughout the model and use Xavier initialization (Glorot and Bengio, 2010) for the parameters. The model is trained with ADAM optimizer (Kingma and Ba, 2014) on L2-regularized negative log-likelihood. The regularization strength was set to 0.01, and the dropout was set to 0.35. The best hyper-parameters for the number of basis classifiers is $K = 3$ for the follow and mention networks, and $K = 4$ for the retweet network.

| System | Accuracy |
|---|---|
| Owoputi et. al. | **92.80%** |
| BiLSTM tagger | 90.50% |
| Ensemble of BiLSTM taggers | 90.11% |
| BiLSTM taggers with social attention | 89.80% |

Table 4: Accuracy of the models on the DAILY547 dataset. The best results are in **bold**.

| Network | Accuracy |
|---|---|
| Follow | 89.42% |
| Mention | 89.80% |
| Retweet | 89.65% |

Table 5: Accuracy of the social attention model, across each of the three networks.

### 5.2 Results and Discussion

Table 4 summarizes the main empirical findings, where we report results from author embeddings trained on the mention network for Social Attention. The results of different social networks with Social Attention is shown in Table 5.

We also evaluate the performance of the trained Social Attention model on the subset of authors who can be located in the social network. The accuracy on these authors is similar to the overall performance on the full dataset. We also observe the attention distributions of the authors in the social network on the basis models in the ensemble. For every pair of authors $a_i$ and $a_j$ connected in the social network we compute $\Sigma_k |\pi_{a_i,k} - \pi_{a_j,k}|$ and average it across all pairs in the network. This is compared with against a randomly rewired network. If this value is lower for the social network, then this indicates that the connected authors tend to have similar attentional distributions as explained in § 4. The results are presented in Table 6. These results clearly indicate that the authors who are connected in the social network tend to have similar attentional distributions.

| Network | Actual Network | Random |
|---|---|---|
| Follow | 0.90 | 1.10 |
| Mention | 0.38 | 1.06 |
| Retweet | 0.36 | 0.68 |

Table 6: Comparison of the mean absolute difference in attention distributions of connected authors in actual social networks versus randomly rewired networks.

While the analyses in § 3 indicated a strong degree of linguistic homophily, we do not observe any significant gain in performance. We think the following factors played an important role:

**Missing authors.** There are a large number of missing authors in each of the social network (about 50% of the authors of the tweets in the dataset). The results from combining all the three social networks by just concatenating this embeddings did not help either in our experiments.

**Tweets per author.** We have only one tweet for every author in our dataset and this makes it harder for the model to extract relations between authors and their tweets.

**Dataset size.** The dataset contains only 2374 tweets, which could be the reason our deep learning model is still behind the feature-rich Markov Model of Owoputi et al. (2013) by about 2%.

**Sparse social networks.** The social networks that we constructed using the twitter IDs from the tweet metadata of the OCT27 and DAILY547 datasets were very sparse, and the node degree distributions (number of edges per node) have high variance.

## 6 Related Work

Previous problems on incorporating social relations have focused on sentiment analysis and entity linking, where the existence of social relations between users is considered as a clue that the sentiment polarities in the messages from the users should be similar or the entities that they refer to in their messages are the same. Speriosu et al. (2011) constructs a heterogeneous network with tweets, users, and n-grams as nodes, and the sentiment label distributions associated with the nodes are refined by performing label propagation over social relations. Tan et al. (2011) and Hu et al. (2013) leverage social relations for sentiment analysis by exploiting a factor graph model and the graph Laplacian technique respectively, so that the tweets belonging to social connected users share similar label distributions. Yang et al. (2016) proposed a neural based structured learning architecture for tweet entity linking, leveraging the tendency of socially linked individuals to share similar interests on named entities — the phenomenon of entity homophily. Yang and Eisenstein (2017) proposed a middle ground between group-level demographic characteristics and personalization, by exploiting social network structure. We extend this work by applying it for the first time to syntactic analysis.

## 7 Conclusion

This paper describes the hypothesis of linguistic homophily specifically linked to stylistic variation on social media data and tests the effectiveness of social attention to overcome language variation, leveraging the tendency of socially proximate individuals to use language similarly for POS tagging. While our preliminary analyses demonstrate a strong correlation between tagging accuracy and network structure, we are unable to leverage these correlations for improvements in tagging accuracy.

How should we reconcile these conflicting results? In the limit of infinite resources, we could train separate taggers for separate treebanks, featuring each language variety. But even if language variation is strongly associated with the network structure, the effectiveness of this approach would still be limited by the inherent difficulty of tagging each language variety. In other words, augmenting the tagger with social network metadata may not help much, because some parts of the network may simply be harder to tag than others. However, this pessimistic conclusion must be offset by noting the small size of existing annotated datasets for social media writing, which are orders of magnitude smaller than comparable corpora of newstext. While some online varieties maybe hard to tag well, it is equally possible that the advantages of more flexible modeling frameworks only become visible when there is sufficient data to accurately estimate them. We are particularly interested to explore the utility of semi-supervised techniques for training such models in future work.